\documentclass[nohyperref]{article}

\usepackage{microtype}
\usepackage{graphicx}
\usepackage{subfigure}
\usepackage{booktabs} 

\usepackage{hyperref}

\usepackage[accepted]{icml2022}

\usepackage{amsmath}
\usepackage{amssymb}
\usepackage{mathtools}
\usepackage{amsthm}

\usepackage[capitalize,noabbrev]{cleveref}

\theoremstyle{plain}

\theoremstyle{definition}

\theoremstyle{remark}

\usepackage[textsize=tiny]{todonotes}

\icmltitlerunning{Time Series Anomaly Detection by Cumulative Radon Features}

\begin{document}

\twocolumn[
\icmltitle{Time Series Anomaly Detection by Cumulative Radon Features}



\icmlsetsymbol{equal}{*}

\begin{icmlauthorlist}
\icmlauthor{Yedid Hoshen}{yyy}
\end{icmlauthorlist}

\icmlaffiliation{yyy}{School of Computer Science and Engineering, The Hebrew University of Jerusalem, Israel}

\icmlcorrespondingauthor{}{}

\icmlkeywords{Anomaly detection, Time series}

\vskip 0.3in
]



\printAffiliationsAndNotice{}  

\begin{abstract}
Detecting anomalous time series is key for scientific, medical and industrial tasks, but is challenging due to its inherent unsupervised nature. In recent years, progress has been made on this task by learning increasingly more complex features, often using deep neural networks. In this work, we argue that shallow features suffice when combined with distribution distance measures. Our approach models each time series as a high dimensional empirical distribution of features, where each time-point constitutes a single sample. Modeling the distance between a test time series and the normal training set therefore requires efficiently measuring the distance between multivariate probability distributions. We show that by parameterizing each time series using cumulative Radon features, we are able to efficiently and effectively model the distribution of normal time series. Our theoretically grounded but simple-to-implement approach is evaluated on multiple datasets and shown to achieve better results than established, classical methods as well as complex, state-of-the-art deep learning methods. Code available at \url{https://github.com/yedidh/radonomaly}
\end{abstract}

\section{Introduction}

Detecting anomalies in temporal sequences of observations is a key aspect of computational and artificial intelligence. The objective of time series anomaly detection is to determine if a temporal sequence contains patterns that are semantically different from those previously seen. This has important applications in science (e.g. detecting unusual stellar orbits for discovering black holes), medicine (e.g. detecting unusual ECG patterns) and industry (e.g. detecting unusual network traffic patterns for intrusion detection). Despite the importance of the task and the significant research effort spent on solving it, it remains very challenging. The main aspect that makes the anomaly detection task difficult in general is the limited supervision. Anomalies typically cannot be assumed to have been seen in training as they are by nature rare and unexpected, a successful anomaly detector must deal with all anomalies despite not seeing them previously.  

Density estimation is a successful paradigm for anomaly detection. The idea is to first estimate the probability density of the normal data at training time. At test time, the probability density of the target sample is evaluated - samples lying in low-density regions are designated as anomalous. Although conceptually simple, estimating the probability density of high-dimensional data is very hard as the distributions of real-world data are complex while limited dataset sizes reduce the estimation quality. Time series anomaly detection is particularly hard as the task requires understanding complex, non-stationary temporal patterns. Previous time series anomaly detection approaches use different techniques for overcoming this issue. Several notable lines of work include: i) parametric density estimation of the normal data iii) non-parametric density estimators (e.g. K nearest neighbors ) iii) using the estimation errors of auto-regressive models iv) evaluating the performance of classifiers trained on auxiliary tasks. In recent years, deep learning methods have been used to significantly improve each of the above techniques by learning better features. None-the-less, current time series anomaly detection accuracy is insufficient for many real-world uses.

In Sec.~\ref{sec:method}, we tackle the task of time series density estimation by treating each time series as a high-dimensional multivariate probability density function (PDF). The features of each time-point along the series are treated as a sample from the PDF. We estimate the PDF by the empirical PDF, which is very sparse. The goal is to estimate the distribution of all normal time series, which is essentially estimating the distribution over the PDFs, each representing a normal time series. This goal can be made more achievable by representing each PDF using a compact representation. Our method is to compute multiple projections of the high-dimensional PDF using the Radon transform (Sec.~\ref{subsec:radon_bg}). Each projection is a far denser PDF than the original multivariate PDF. We estimate the cumulative probability distribution of each projection using a histogram estimator. Each time series is therefore now represented as a set of histograms, which we denote, the cumulative Radon features. We evaluate a number of distance measures and density estimators over the cumulative Radon features. We highlight that simple distance measures over the cumulative Radon features are closely related to the Sliced Wasserstein Distance. Given the density estimators, we combine two time series anomaly scoring methods with the cumulative Radon features: distance from the center and distance from the K nearest neighbors (kNN). 
As the cumulative Radon features are highly correlated, standard distance measures in the original feature space assign disparate importance to different factors of variations. Instead, we suggest sphering the data to remove this correlation. Our final features, the \textit{sphered cumulative Radon features} -  are effective with both anomaly scoring methods.

A thorough experimental investigation of our method is presented in Sec.~\ref{sec:evaluation}. Multiple datasets are used spanning the range of practical time series anomaly detection applications. Our method was evaluated against deep learning methods recently published in top conferences, as well as established, classical methods. Our method typically performed better than all previous methods despite being easy-to-implement and having short inference times. Finally, the limitations of our work and possible future directions are discussed in Sec.~\ref{sec:discussion}.  

\section{Previous work}

\textbf{Time series Anomaly detection.} The task of anomaly detection in time series has been studied over several decades, see \citet{blazquez2021review} for a comprehensive survey. In this paper, we are mostly concerned with collective anomaly detection i.e. abnormal patterns in a collection of points. Traditional approaches for this task include generic anomaly detection approaches such as: K nearest neighbors (kNN) based methods e.g. vanilla kNN \citep{eskin2002geometric} and Local Outlier Factor (LOF) \citep{breunig2000lof}. Tree-based methods e.g. Isolation Forest \citep{liu2008isolation}. One-class classification methods e.g. One-Class SVM \citep{tax2004support} and SVDD \citep{scholkopf1999support}. Some traditional methods are particular to time series anomaly detection, specifically auto-regressive methods \citep{rousseeuw2005robust}. With the advent of deep learning, the traditional approaches were augmented with deep-learned features. Deep one-class classification methods include DeepSVDD \citep{ruff2018deep} and DROCC \citep{goyal2020drocc}. Deep auto-regressive methods include RNN-based prediction and auto-encoding methods e.g. \citet{bontemps2016collective} and \citet{malhotra2016lstm}. In addition, some deep learning anomaly detection approaches were proposed that are conceptually different from traditional approaches. These methods are based on the premise that classifiers trained on the normal data will struggle to generalize to anomalous data. These approaches were originally developed for image anomaly detection \citep{golan2018deep} but have been extended to tabular and time series data e.g. by \citet{bergman2020classification} and \citet{qiu2021neural}.

\textbf{Radon transform.} First introduced by \citet{radon20051}, it transforms a signal to its projections along a set of lines. Although originally introduced for the 2D and 3D functions, it has been extended to arbitrary dimension \citep{deans2007radon}. The Radon and inverse Radon transform have been applied to many important tasks including medical tomography \citep{quinto2006introduction}. The Radon and Fourier transforms are related through the Projection-Slice theorem.   

\textbf{Projection-based multivariate probability representation.} Much research has been done on representing and estimating high-dimensional multivariate probability distributions. One line of work proposed to represent them by projections to a single dimension. The motivation is that PDF estimation of a RV in $\mathbb{R}$ is usually easier than in $\mathbb{R}^d$, where $d$ is a large number. In anomaly detection, HBOS \citep{goldstein2012histogram} and LODA \citep{pevny2016loda}, represent the distribution of normal data by multiple 1D projections. Note that differently from our method, those approaches model each sample as a single point whereas we treat every time series as a set of samples, leading to significant differences. \citet{kolouri2015radon} represent images using the CDF-transform of Radon projections and show that these features provide an effective non-linear basis for classification. Differently from them, we deal with time series and our objective is anomaly detection rather than supervised classification. Also note that the CDF-transform is the Sliced Wsserstein-2 distance, while we consider a larger class of distance functions. It has also been shown by \citet{bonneel2015sliced} that projections can be used to efficiently estimate the Wasserstein barycenter of a distribution. The connection between the Radon transform and the Sliced Wasserstein Distance has attracted a line of work attempting to find an alternative to the difficult optimization of Wasserstein GAN \citep{arjovsky2017wasserstein}, using the easy-to-compute  Sliced Wasserstein Distance e.g. \citet{deshpande2018generative} and \citet{kolouri2019generalized}. Our work is able to directly connect this line of research to the task of time series anomaly detection. Finally, Rocket  \citep{dempster2020rocket} and its followup work MiniRocket \citep{dempster2021minirocket} proposed projection-based bag-of-words features which were empirically very effective for time series classification. Our work focuses on anomaly detection rather than classification, and is theoretically motivated by the Radon-transform. These works can be seen as a special case of our formulation.

\section{Method}
\label{sec:method}

Our approach models a time series as a set of independent samples from a multivariate probability density function. Modeling the distribution of normal data requires an efficient method for modeling the distribution of multivariate PDFs. We encode each PDF efficiently by transforming it using the Radon transform. This allows us to utilize powerful probability distance measures for measuring the distance between test samples and the normal data. 

\subsection{Preliminaries}

We denote each time series $S = [X_1..X_T]$, where $X_t$ denotes the point at time $t$. The time series is typically multivariate, $X_t \in \mathbb{R}^d$ where $d$ is the number of dimensions of the time series. The input is a training set of $N_{train}$ time series, all of which are normal. At test time, a new time series $S_{test}$ is presented and our task is to determine if it is normal or anomalous.   

\subsection{A Time series as a Probability Distribution}
\label{subsec:pdf}

We model each time series $S$ as a set of time points $X_t$. To take the context of the point $X_t$ into account, we represent it using a temporal window of the surrounding points $X_{t-w}..X_t..X_{t+w}$. Following previous work \citep{dieleman2013multiscale, dempster2020rocket}, we include windows of $N_r$ different resolutions (we use as many resolutions as allowed by length $T$). As the size of the window increases linearly with the resolution, some form of subsampling is required, we subsample all windows to $2w + 1$ samples with a fixed stride. Unless otherwise specified, the representation simply uses the concatenation of the raw values of the points included in the temporal windows. The $d$ dimensions of the points are also combined by concatenation. Note that this feature representation is quite naive, but we find that even such a simple representation can already achieve strong anomaly detection performance. At the end of this stage, each point $X_t$ is represented by a vector of dimension $d_f = (2*w + 1) \times N_r \times d$. We denote this feature vector of point $X_t$ as $f_t \in \mathbb{R}^{d_f}$. The time series is represented by the set $F = \{f_1,,f_T\}$ containing the features of all its points.

The set of points is now encoded as an empirical probability distribution function $P(f)$ where $P:\mathbb{R}^{d_f} \rightarrow \mathbb{R}^{+}$. We define $P$ such that:

\begin{equation}
P(f) = \sum_i \frac{1}{T} \delta(f - f_i)
\end{equation}

Where $\delta$ is a Dirac delta function. Note that $P$ is very sparse as it is estimated from a small number of samples (due to length of the time series) while being defined over a high-dimensional space.

\subsection{Compactifying the Representation}

Representing the time series by the empirical PDF $P$ has presented two inter-related questions: i) How to measure the distance between the PDFs of two time series? ii) How to represent the distribution of PDFs of the entire set of normal time series? To answer both questions, we require a compact representation of the PDF. This will be described in this section.

\subsubsection{Background: the Radon Transform}
\label{subsec:radon_bg}

The Radon transform projects a PDF along a set of directions $\theta \in S^{d_f - 1}$. 

\begin{equation}
  \tilde{P}(\theta, t) = \int_{f \in \mathbb{R}^{d_f}} P(f) \delta(t - \langle \theta, f \rangle) df 
\end{equation}

Although this integral is expensive to compute in the general case, here it can be computed in closed form as our PDF consists of a sum of Dirac delta functions. 

Each one-dimensional (1D) projection of the PDF is far more dense than the original PDF. On the other hand, it only describes a single projection of the PDF. An important aspect of the Radon transform, is that given the entire set of 1D projections, it is possible to invert the transformation and recover the original PDF: 

\begin{equation}
\label{eq:inv_radon}
  P(f) = \int_{\theta \in \mathbb{S}^{d_f - 1}} \tilde{P}(\theta, \langle \theta , f \rangle) * \eta(\langle \theta, f \rangle)) d\theta 
\end{equation}

Where $*$ denotes the convolution operation and $\eta$ is a particular high-pass filter (see \citet{kolouri2015radon} for details). The Radon transform therefore does not result in a loss of information. Note that exact recovery of $P$ is generally possible only with a very large number of projections $\theta$, however good approximations may be obtained with a much smaller number of projections.

The projection along each direction $\tilde{P}(\theta, .)$ is a valid probability density function. We therefore denote the cumulative distribution function $J$:  

\begin{equation}
  J(\theta, t) = \int_{t'=-\infty}^{t} \tilde{P}(\theta, t') dt'
\end{equation}

It is clear that $J(\theta, -\infty) = 0$ and $J(\theta, \infty) = 1$. 

\subsubsection{Radon Features}
\label{subsec:radonm}

The Radon transform presents advantages and disadvantages for measuring the distance between the probability distributions of two time series. On the one hand, the projected PDFs along each direction $\theta$ are far denser than in the original space $f$, making it amenable to different distance measures (that will be described in Sec.~\ref{subsec:ablations}). On the other hand, the Radon transform projects $P$ along an infinite number of directions $\theta$, which is impractical. To overcome this limitation, we approximate the entire set of projections by subsampling a set of directions $\Omega$. Note that choosing $\Omega$ as the set of standard bases (i.e. along the first dimension, the second dimension etc.), the approximate Radon transform becomes identical to the marginals of the PDF along the raw axes. However this may provide a biased estimate of $P$, i.e. inverting this subset of Radon projections using Eq.~\ref{eq:inv_radon} is unlikely to recover the original $P$ . We evaluate the effects of the choice of $\Omega$ in Sec.~\ref{tab:ablation}. Having selected a the set of Radon projections $\{ \tilde{P}(\theta, t) | \theta \in \Omega \}$, we estimate the cumulative probability density of each projection $J(\theta, t)$ using a histogram estimator. To conclude, each time series is described by the set of values of the cumulative density functions $J(\theta, t)$  along a set of directions $\theta \in \Omega$. We denote this set of features of time series $S$ as $\mathcal{CR}(S)$. Assuming that the number of projection directions is $N_{P}$ (we use $N_P=100$ unless stated otherwise) and number of bins in the histogram is $N_B$ (we use $N_B=20$ unless stated otherwise), each time series is now represented in $N_P \times N_B$ dimensions.

\subsection{Distance Measures}
\label{subsec:distance}

We define different distance measures between time series, using the cumulative Radon features described in Sec.~\ref{subsec:radonm}.

\textbf{Sliced Wasserstein Distance.} Optimal transport is one of the most popular ways of measuring the distance between probability distributions (and sets in general). The Wasserstein (or Earth Mover's Distance) measures the minimal distance needed for the distribution to be transformed from one probability distribution to the other. As computing the Wasserstein distance for high-dimensional data such as ours is computationally demanding, the Sliced Wasserstein Distance (SWD) was proposed as an alternative. The SWD was theoretically shown to have many attractive properties e.g. \citet{bonneel2015sliced}. Both the $SWD_1$ and $SWD_2$ distances can be easily expressed using the Radon features. The $SWD_1$ has a particularly simple form:

\begin{equation}
  SWD_1(S_1, S_2) = \sum_t \sum_{\theta} \|\mathcal{CR}(S_1) - \mathcal{CR}(S_2) \|_1
\end{equation}

The $SWD_2$ has a somewhat more complex formulation to the Radon features. We follow the formulation of the Radon-CDF transform \cite{kolouri2015radon} in defining the threshold $s(\theta, t)$ where $J_1(\theta, s(\theta, t)) = J_2(\theta, t)$. Using this notation the $SWD_2$ can be written as:

\begin{equation}
  SWD_2(S_1, S_2) = \sqrt{\sum_t \sum_{\theta} \| s(\theta, t) - t\|_2^2}
\end{equation}

\textbf{Euclidean Cumulative Radon Distance:.} The $SWD_1$ can be seen to be the $L_1$ distance between the cumulative Radon features of the two time series. We also explore the $L_2$ distance between Cumulative Radon features.  

\begin{equation}
  L_2(S_1, S_2) =  \| \mathcal{CR}(S_1) - \mathcal{CR}(S_2)\|_2^2
\end{equation}

\subsection{Sphered Radon Features}
\label{subsec:decorr}

The distances described in Sec.~\ref{subsec:distance} are data agnostic i.e. they do not use the distribution of normal time series, but are rather defined apriori. In fact, the Radon features are strongly correlated as different projections of the same data have strong correlations. Additionally, the bins in the individual histograms are strongly correlated (this is obvious for 1D-CDFs but it is also true for the PDF). We propose a simple, but also highly effective solution, namely, sphering the Radon features. Specifically we use ZCA sphering. We compute the mean (denoted by $\mu$) and the covariance $\Sigma$ of the features of all training time series (which are all normal in this setting). We then transform the features of all time series by mean subtraction, transformation to the principle axes, and sphering of the spectrum:

\begin{equation}
    \mathcal{SR}(S) = \Sigma^{-\frac{1}{2}}(\mathcal{CR}(S) - \mu)
\end{equation} 

After sphering, the features of the normal time series have an identity covariance matrix. In Sec.~\ref{sec:evaluation} we demonstrate that sphered Radon features outperform all other approaches for time series anomaly detection.

\subsection{Anomaly Scoring}

We evaluate different distance-based criteria for classifying a test time series as normal or anomalous.

\textbf{Distance to mean:} We evaluate the distance between the test time series and the mean of the features of the normal time series. 

\textbf{K nearest neighbors (kNN):} We compute the distance between the features of the test time series and every time series in the normal training set. The average distance to the K nearest time series is used as the anomaly criterion.

\section{Experiments}
\label{sec:evaluation}

In this section, we extensively evaluate our approach against a large range of time series anomaly detection approaches. Our evaluation spans both long standing, time-tested methods as well as more recent deep learning methods that currently achieve the state-of-the-art on different benchmark datasets. In Sec.~\ref{subsec:synth}, we characterize the performance of our method using a synthetic dataset which has carefully crafted anomalies of different types. In Sec.~\ref{subsec:real}, we evaluate our method against a large number of methods on a set of benchmark anomaly detection datasets. In Sec.~\ref{subsec:traj}, our method is evaluated on activity anomaly detection on landmark trajectory data against specialized deep learning approaches. We perform extensive ablations in Sec.~\ref{subsec:ablations}.

\subsection{Characterizing the Detectable Anomalies}
\label{subsec:synth}

\begin{table*}
\caption{Results on the TODS dataset. Average ROCAUC over the $4$ anomaly ratios (top two methods in \textbf{bold} and \underline{underline}).}
\centering
\begin{tabular}{lcccccc}
\toprule

	&	\textbf{Shapelet}	&	\textbf{Trend}	&	\textbf{Seasonal}	&	 \textbf{Point Context}	&	\textbf{Point Global}	&	\textbf{Average}	\\ \midrule
AR	&	0.56	&	0.5	&	0.335	&	\textbf{0.6775}	&	0.9125	&	0.597	\\
GBRT	&	\underline{0.675}	&	0.505	&	0.61	&	0.5375	&	0.7875	&	\underline{0.623}	\\
LSTM-RNN	&	0.14	&	0.5425	&	0.265	&	0.085	&	0.29	&	0.2645	\\
IForest	&	0.0075	&	0.1075	&	0.18	&	0.2425	&	0.89	&	0.2855	\\
OCSVM	&	0.04	&	0.1375	&	0.1375	&	0.2875	&	\underline{0.925}	&	0.3055	\\
AutoEncoder	&	0.065	&	0.12	&	0.05	&	0.0875	&	0.8225	&	0.229	\\
GAN	&	0.065	&	0.12	&	0	&	0.0875	&	0.5425	&	0.163	\\
NIForest	&	0.3175	&	0.595	&	0.37	&	0.23	&	0.225	&	0.3475	\\
NOCSVM	&	0.42	&	\underline{0.6025}	&	\textbf{0.495}	&	0.23	&	0.2275	&	0.395	\\
MatrixProile	&	0.5125	&	0.5725	&	0.3175	&	0.155	&	0.165	&	0.3445	\\
Ours	&	\textbf{0.735}	&	\textbf{0.731}	&	\underline{0.467}	&	\underline{0.622}	&	\textbf{0.934}	&	\textbf{0.6978}	\\

\bottomrule
\end{tabular}
\label{tab:tods}
\vspace{5pt}
\end{table*}

In this section, we investigate the characteristics of the anomalies detectable by our model. To this end, we use a synthetic dataset \citep{lai2021revisiting} specifically designed for this analysis. We use the univariate evaluation suite which consists of $20$ different settings. In each setting, the normal data consists of a sinusoidal wave of length $200$. Anomalies are injected at random locations. There are five different types of anomalies: three are group anomalies - Trend, Shapelet, and Seasonal, two are point anomalies: Global and Contextual. Each scenario is presented in four different anomaly ratios: $5\%$, $10\%$, $15\%$, $20\%$. For each time point, the extracted features are simply the surrounding $9$ time-points. We compute sphered cumulative Radon features. We score anomalies using the Euclidean distance from the center. This approach performs very well for collective anomalies - but is less well suited for point anomalies (As the features are extracted for the whole window rather than a single point). Therefore, for the point anomaly scenarios, we extract cumulative Radon features for the window - and train a linear regressor on these features to predict the value of the next time point. The anomaly score is the  error between the predicted and actual value of the point.

\textbf{Results.} The comparison between all methods is presented in Tab.~\ref{tab:tods}. We can see that our method performs particularly well for collective anomalies. In the shapelet and trend scenarios, it performs the next best method by a significant margin. On point anomalies, it performs comparably to auto-regressive models. This is expected, as our variant to point anomalies is indeed a type of AR model. The strong performance of our method is particularly remarkable as no method performs well in all settings, whereas we see that cumulative Radon features are indeed powerful for all anomaly scenarios (but note that different scoring methods were needed in the group and point scenarios).

\subsection{Evaluation on Realistic Benchmark Datasets}
\label{subsec:real}

Have established that our method is particularly powerful for collective anomalies - we now evaluate it on more realistic multivariate time series anomaly detection datasets. In this section, we follow the experimental protocol of  \citet{qiu2021neural}. We evaluate our method on $5$ different datasets taken from timeseriesclassification.com. The datasets were originally designed for multi-class classification. They were converted to anomaly detection datasets, by designating a single class as normal and all other classes as anomalies. The models are trained on the training set of just the normal data. The entire test set (normal and anomalous) is used for evaluation. This procedure is repeated by creating multiple datasets, each designating a different class as normal. The final accuracy is the average of the ROCAUC for each of these datasets. All the results of the baselines were copied from \cite{qiu2021neural}. The datasets are:

\textit{RacketSports (RS).} Accelerometer and gyroscope recording of players playing four different racket sports. Each sport is designated as a different class.

\textit{Epilepsy (EPSY).} Accelerometer recording of healthy actors simulating four different activity classes, one of them being am epileptic shock.

\textit{Naval air training and operating procedures standardization (NAT).} Positions of sensors mounted on different body parts of a person performing activities. There are six different activity classes in the dataset.

\textit{Character trajectories (CT).} Velocity trajectories of a pen on a WACOM tablet, with $20$ different characters classes. 

\textit{Spoken Arabic Digits (SAD).} MFCC features of ten arabic digits spoken by $88$ different speakers. We follow the processing of the dataset as done by \citet{qiu2021neural}. In private communications the authors explained that only sequences of lengths between 20 and 50 time steps were selected. The other time series were dropped.

\begin{table*}
\caption{Performance comparison on realistic datasets. Average ROCAUC over all classes.}
\centering
\small
\begin{tabular}{lccccccccccc}
\toprule

	&	OCSVM	&	IF	&	LOF	&	RNN			&	LSTM-ED			&	DeepSVDD			&	GOAD			&	DROCC			&	NeuTraL			&	Ours			\\ \midrule
EPSY	&	61.1	&	67.7	&	56.1	&	80.4	$\pm$	1.8	&	82.6	$\pm$	1.7	&	57.6	$\pm$	0.7	&	76.7	$\pm$	0.4	&	85.8	$\pm$	2.1	&	92.6	$\pm$	1.7	&	\textbf{98.1}	$\pm$	0.3	\\
NAT	&	86	&	85.4	&	89.2	&	89.5	$\pm$	0.4	&	91.5	$\pm$	0.3	&	88.6	$\pm$	0.8	&	87.1	$\pm$	1.1	&	87.2	$\pm$	1.4	&	94.5	$\pm$	0.8	&	\textbf{96.1}	$\pm$	0.1	\\
SAD	&	95.3	&	88.2	&	98.3	&	81.5	$\pm$	0.4	&	93.1	$\pm$	0.5	&	86	$\pm$	0.1	&	94.7	$\pm$	0.1	&	85.8	$\pm$	0.8	&	\textbf{98.9}	$\pm$	0.1	&	97.8	$\pm$	0.1	\\
CT	&	97.4	&	94.3	&	97.8	&	96.3	$\pm$	0.2	&	79	$\pm$	1.1	&	95.7	$\pm$	0.5	&	97.7	$\pm$	0.1	&	95.3	$\pm$	0.3	&	99.3	$\pm$	0.1	&	\textbf{99.7}	$\pm$	0.0	\\
RS	&	70	&	69.3	&	57.4	&	84.7	$\pm$	0.7	&	65.4	$\pm$	2.1	&	77.4	$\pm$	0.7	&	79.9	$\pm$	0.6	&	80	$\pm$	1	&	86.5	$\pm$	0.6	& \textbf{92.3}	$\pm$	0.3	\\ \midrule
Avg.	&	82.0	&	81.0	&	79.8	&	86.5			&	82.3			&	81.1			&	87.2			&	86.8			&	94.4			&	\textbf{96.8	}		\\

\bottomrule
\end{tabular}
\label{tab:realworld}
\end{table*}

Our method was compared to a comprehensive set of methods: One-class classification - One-class SVM (OC-SVM) and its deep versions DeepSVDD (commonly used) and DROCC (recently published) are one-class classifiers. Tree-based - Isolation Forest (IF). Nearest-neighbors - LOF as a specialized version of nearest neighbor anomaly detection. Auto-regressive - RNN and LSTM-ED are deep auto-regressive prediction models and are probably the most commonly used methods in time series anomaly detection. Transformation prediction - GOAD and NeuTraL-AD are based on transformation prediction, and are adaptations of RotNet-based approaches (such as GEOM \citep{golan2018deep}). \citet{qiu2021neural} also compared to the deep auto-encoding method DAGMM \citep{zong2018deep} but as it did not perform the presented method on any of the datasets, we do not present it here.

\textbf{Results.} Our results are presented in Tab.~\ref{tab:realworld}. We can observe that within the baselines, different approaches are effective for different datasets. kNN-based LOF is highly effective for SAD which is a  large dataset but achieves worse results for EPSY. Auto-regressive approaches achieve strong results on CT.  Transformation-prediction approaches, GOAD and NeuTraL are the top performing baselines. The learned transformations of NeuTraL achieved better results than the random transformations of GOAD. Our method achieves the best overall results both on average and individually on all datasets apart from SAD (where it is comparable but a little lower than NeuTraL). Differently from NeuTraL, our method is simple, does not use deep neural networks, is very fast to train and evaluate, has few hyper-parameters and is theoretically justified.

\subsection{Anomalous Activity Detection by Landmark Trajectories }
\label{subsec:traj}

To evaluate our approach on a specific task on which previous approaches were significantly tuned, we consider anomalous activity detection, based on landmark trajectories. As our objective is to evaluate multivariate time series anomaly detection. we only compare to methods that use landmark trajectories, rather than pixel-level features (which can achieve higher accuracy). We follow the finegrained setting of Markovitz et al. \cite{markovitz2020graph}. We use the  ShanghaiTech Campus dataset, a diverse video anomaly detection dataset containing $130$ different anomalous events observed by cameras with different poses and viewing conditions. The dataset consists of $300$ training and $100$ test videos, which are uncurated. Each frame is first processed by a 2D landmark detector for each person individually. Sliding windows of length $12$ are used for viewing the trajectories of the landmarks of each person over $12$ frames. The training set consists of only normal trajectories, while the test set contains a mixture of normal and anomalous trajectories. A trajectory anomaly detector is trained using the normal training set. Note, that the detector only observes the trajectory of a single person. If a frame contains multiple persons, the score of the most anomalous person is used for this frame. Finally, the ROCAUC is computed over all the frames in the test set. In practice, we used the authors' code both for preprocessing the trajectories and for scoring the results. We compare using our method instead of the anomaly detection method proposed by \citet{markovitz2020graph}. Their method, named ST-GCAE, is based on clustering the embeddings of graph convolutional neural networks. It consists of both a pretraining and a finetuning stage. Our method in comparison simply consists of extracting cumulative Radon features and ZCA-sphering using their covariance matrix. 

\textbf{Results.} The results of the evaluation are presented in Tab.~\ref{tab:trajectory}. Our method achieves better results than the much more complex and compute intensive state-of-the-art approaches of \citet{morais2019learning} and \citet{markovitz2020graph}. We repeat that we do not claim that this is the state-of-the-art on the ShanghaiTech Campus dataset as using pretrained networks on pixel data can achieve  higher accuracy. We only claim our method achieves better results than other methods that use similar landmark trajectory data.    

\begin{table}
\caption{Anomaly detection accuracy using landmark trajectory data on the ShanghaiTech Campus Dataset (ROCAUC).}
\vspace{5pt}
\centering
\begin{tabular}{lc}
\toprule

Method	&	Accuracy\\ \midrule
\citet{luo2017revisit}	&	0.68\\
\citet{abati2019latent}	&	0.725\\
\cite{liu2018future}	&	0.728\\
\cite{morais2019learning}	&	0.734\\
\cite{markovitz2020graph}	&	0.752\\
Ours	&	0.761 $\pm$ 0.3		\\

\bottomrule
\end{tabular}
\label{tab:trajectory}
\end{table}

\subsection{Ablations}
\label{subsec:ablations}

This section provides an extensive ablation of our method.

\begin{figure*}
  \centering
  \includegraphics[width=0.4\linewidth]{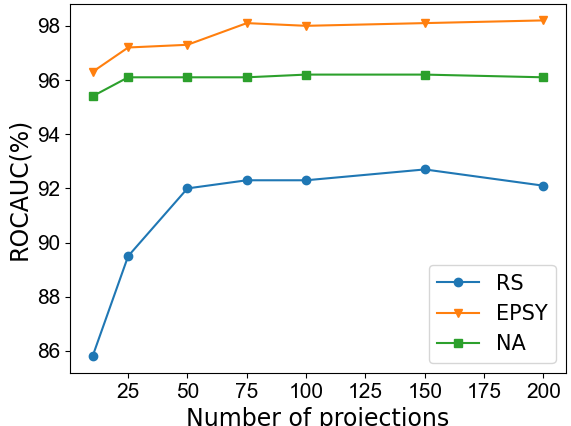}~~~
  \includegraphics[width=0.4\linewidth]{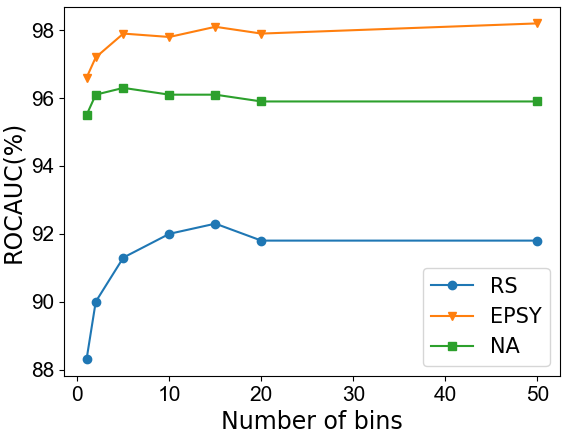}
  \caption{Ablation of our method's accuracy vs. number of projections (left) and number of bins (right). }
  \label{fig:slice_bin}
  \vspace{5pt}
\end{figure*}

\textbf{Number of projections.} The Radon transform requires a large number of projections to retain all information in the original PDF. However, using more projections increases the computational load and runtime of our method. We investigate the effect of the number of projections over the final accuracy of our method. The results are provided in Fig.~\ref{fig:slice_bin}. We can observe that although a small number of projections hurts performance, even a moderate number of projections is sufficient. We therefore use the value of $100$ projections in our experiments unless otherwise specified.

\textbf{Number of bins.} Our method requires approximating the CDF (or PDF) of the distribution along each projection. We use a histogram density estimator which approximates the CDF/PDF. Although a perfect approximation would require storing the same number of values as the number of data points, in practice we use a smaller number of bins. In Fig.~\ref{fig:slice_bin}, we present the accuracy of our method on three datasets as a function of the number of bins (per-projection). The experiments were run with 100 projections. Our results show that beyond a very small number of bins - larger numbers are not critical. We use $20$ bins which we empirically found to be sufficient in all our experiments, while allowing for fast computation (particularly of spectral decomposition of the covariance matrix) and a low memory footprint.

\textbf{Effect of sphering.} Standard probability distances such as the sliced Wasserstein-1 and Wasserstein-2 distances are not learned. They do not estimate the correlation between different bins in the histogram and therefore do not necessarily assign equal importance to all degrees of variation. The Mahalanobis distance uses the covariance matrix of the data across a training distribution and is therefore able to reason about the components of variation. On the other hand, the Mahalanobis distance is between a sample and a distribution of samples, rather than a distribution and a set of distributions. Here, we proposed a way to utilize the Mahalanobis distance to estimate the distribution of distributions by sphering the Radon features. The raw sphered features ("sph.") are presented in Tab.~\ref{tab:ablation}. The results demonstrate that sphering is a crucial component in our approach.  

\textbf{Comparing distance functions for kNN.} An alternative to the distance from the center is using kNN between the test sample and all the normal training data. kNN critically depends on the distance measure. Here, we ablate several distribution distances: the SWD-1 and SWD-2 (computed exactly as in \cite{kolouri2019generalized}). and $L_1$ and $L_2$ between the CDF of the projections. Note that L1 of the CDF and SWD-1 are equivalent up to quantization. The results are presented in Tab.~\ref{tab:ablation}. We observe that $L_1$ and SWD-1 achieve superior results on EPSY and RS, while results are comparable on the other datasets.

\begin{table}
\caption{Ablation results (ROCAUC).}
\centering
\begin{tabular}{lccccc}
\toprule

	&	EPSY	&	RS	&	NA	&	CT	&	SAD	\\ \midrule
kNN-$L_1$	&	97.7	&	89.1	&	95.2	&	99.5	&	94.2	\\
kNN-$L_2$	&	95.5	&	86.3	&	95.5	&	99.5	&	94.5	\\
kNN-SWD-1	&	95.7	&	86.5	&	95.3	&	99.5	&	94.3	\\
kNN SWD-2	&	96.2	&	88.3	&	95.3	&	99.5	&	94.0	\\ 
Sph. kNN-$L_1$ &	98.3	&	91.9	&	95.8	&	99.7	&	97.6	\\ 
Sph. kNN-$L_2$ &	97.9	&	92.3	&	95.9	&	99.7	&	97.7	\\ 
\midrule
Raw. $L_1$	&	65.7	&	70.6	&	93.3	&	98.6	&	79.1 \\
Raw. $L_2$	&	62.1	&	70.9	&	93.6	&	98.5	&	78.8 \\
Sph. $L_1$	&	98.5	&	92.0	&	96.0	&	99.7	&	97.8	\\
Sph. $L_2$	&	98.1	&	92.3	&	96.1	&	99.7	&	97.8	\\ \midrule
Id.	&	97.1	&	90.2	&	91.8	&	98.2	&	78.3	\\
PCA	&	98.2	&	91.6	&	95.8	&	99.7	&	96.7	\\
Rand	&	98.1	&	92.3	&	96.1	&	99.7	&	97.8	\\

\bottomrule
\end{tabular}
\label{tab:ablation}
\end{table}

\textbf{Comparing different sphered distances.} We investigated if the Euclidean distance is indeed necessary in combination with the sphered features. We evaluated the $\mathcal{SR}$ features with both $L_1$ and $L_2$ distances. The results were comparable across all datasets. This suggests that the sphering is a key aspect - and the particular norm is less significant.

\textbf{Comparing direction sampling methods.} The choice of sampling method for directions is potentially important for accurate estimation of a high-dimensional PDF with a small number of projections. We present a comparison between three different direction selection procedure: i) Gaussian: random sampling of each dimension of each direction $\theta$ from a random Gaussian ii) marginals: projecting along a particular dimension e.g. $\theta = [0, 0, 1, 0, 0]$. iii) PCA: selecting $\theta$ from the eigenvector of the matrix containing all features $f_i$ from all positions in all training time series. The latter scheme should select the directions with maximum variation but is also computationally expensive. The results are presented in Tab.~\ref{tab:ablation}. We find that the marginals are not as informative as the other approaches (as they provide a biased sampling of the PDF). Suprisingly, we do not see a large difference between PCA and randomly selected directions. As PCA is more computationally intensive than random directions, we use random projections in all experiments unless otherwise specified.     

\section{Discussion}
\label{sec:discussion}

\textbf{Sub-sequence-level anomaly detection.} In this work, we mostly evaluated the setting of whole sequence anomaly detection rather than subsequence anomaly detection (although the ShanghaiTech Campus dataset falls within this setting). We chose not to evaluate this setting as it has been shown that current benchmarks can be dominated by trivial methods. Particularly, \citet{kim2021towards} show that a simple approach of using the $L_2$ norm of the raw features is sufficient for achieving competitive or better than the state-of-the-art on popular datasets such as WADI or SMAP. Furthermore, very strong results using the point-adjust protocol can be achieved using a random baseline. We chose to evaluate on datasets that do not suffer from these limitations.  

\textbf{Incorporating deep features.} It was shown that our method was able to outperform the state-of-the-art in time series anomaly detection without using deep neural networks. Although an interesting and surprising result, given the recent literature, we believe that our approach is complementary to deep features. In fact, our method can be combined with deep feature extractors. One possible way of combining the two, is by using a self-supervised pipeline such as Temporal Neighborhood Coding \citep{tonekaboni2020unsupervised} to learn deep features from raw time series data. This can be used as the input to our method. Another possible way of combining the two is by using an architecture that incorporates Radon features into the pipeline 

\textbf{Anomaly detection beyond time series.} Our method is not restricted to time series and can be applied to any anomaly detection task where the input can be formulated as a set. This may be the case in video anomaly detection. We presented results for landmark trajectories, but the same approach should potentially be applicable to deep features extracted from frame sequences. This can also be applicable to graph anomaly detection applications, although this may require a more significant modification.

\textbf{Limitations.} i) We chose very simple time-point features, to highlight the importance of the Radon transform and sphering. We believe the results can be further improved by incorporating deep features as described above or by more elaborate handcrafted features. ii) Matrix decomposition: our method requires spectral decomposition of the covariance matrix. This prevents the number of the features from being extended indefinitely as the decomposition may take more time than allowed by the operation settings. iii) Number of windows: the quality of the covariance matrix estimation may decrease when the time series do not contain many (uncorrelated) windows.  

\section{Conclusion}
\label{sec:conclusion}

We presented a method for time series anomaly detection using sphered cumulative Radon features. We gave theoretical justification for our method and empirically demonstrated its outstanding performance through extensive experiment. All this was achieved without using deep neural networks. Beyond our specific method, we believe the connection between anomaly detection and Radon features is likely to extend to other modalities that can be formulated as sets. The most immediate future direction is combining our method with deep-learned features.

\bibliography{example_paper}
\bibliographystyle{icml2022}

\end{document}